\documentclass[pdflatex,sn-mathphys-num]{sn-jnl}
\usepackage{longtable}
\usepackage{amssymb}
\usepackage{pifont}
\usepackage{lineno,hyperref}
\modulolinenumbers[5] 
\usepackage{caption}
\usepackage{subcaption}
\usepackage{multicol}
\usepackage{verbatim}
\usepackage{listings}
\usepackage{hyperref}
\hypersetup{colorlinks,allcolors=black}
\usepackage[usenames, dvipsnames]{color}
\usepackage{amsmath}
\usepackage[section]{placeins}
\usepackage{algpseudocode}
\makeatletter
\def\BState{\State\hskip-\ALG@thistlm}
\makeatother
\usepackage{fullpage}
\usepackage{times}
\usepackage{multirow}
\usepackage{fancyhdr,graphicx}
\usepackage{algorithm,algorithmicx,pseudocode}
\usepackage[section]{placeins}
\usepackage{lscape}
\usepackage{longtable}
\usepackage{mathtools}
\usepackage{enumerate}
\usepackage{array}
\usepackage{boondox-cal}
\usepackage{makecell}
\usepackage{float}
\usepackage{multicol}
\usepackage{commath}
\usepackage{multirow}
\include{pythonlisting}
\usepackage{pdflscape}
\usepackage{caption}
\usepackage{booktabs}

\begin{document}
		
\title{Advancing Toward Robust and Scalable Fingerprint Orientation Estimation: From Gradients to Deep Learning}

\author*[1]{\fnm{Amit Kumar} \sur{Trivedi}}
\email{rivedi19@gmail.com}

\author[2]{\fnm{Jasvinder Pal} \sur{Singh}}
 \email{jasvinder.cse@cujammu.ac.in}

\affil[1]{\orgdiv{Department of Computer Science and Engineering}, \orgname{Thapar Institute of Engineering and Technology}, \orgaddress{\street{TIET}, \city{Patiala}, \postcode{147004}, \state{Punjab}, \country{India}}}

\affil[2]{\orgdiv{Department of Computer Science and Engineering}, \orgname{Central University of Jammu}, \orgaddress{\street{Samba}, \city{Jammu}, \postcode{181143}, \state{J\&K}, \country{India}}}

\abstract{The study identifies a clear evolution from traditional methods to more advanced machine learning approaches. Current algorithms face persistent challenges, including degraded image quality, damaged ridge structures, and background noise, which impact performance. To overcome these limitations, future research must focus on developing efficient algorithms with lower computational complexity while maintaining robust performance across varied conditions. Hybrid methods that combine the simplicity and efficiency of gradient-based techniques with the adaptability and robustness of machine learning are particularly promising for advancing fingerprint recognition systems. Fingerprint orientation estimation plays a crucial role in improving the reliability and accuracy of biometric systems. This study highlights the limitations of current approaches and underscores the importance of designing next-generation algorithms that can operate efficiently across diverse application domains. By addressing these challenges, future developments could enhance the scalability, reliability, and applicability of biometric systems, paving the way for broader use in security and identification technologies.}
		
\maketitle
\section{Introduction}
	More than a century ago, \textit{Alphonse Bertillon} conceived the idea of using the measurement of human body to solve the crime related problems \cite{rhodes1956father}, but before gaining any popularity, it was replaced by the more significant and practical distinctiveness of the fingerprint. The \textit{Home Ministry Office of United Kingdom} accepted the individuality of fingerprint in 1893 and used it to identify the person for law enforcement purposes. Many major law enforcement departments started using it to identify the repeated offenders who change their identity after each offense. They started recording the offender fingerprint at the time of arrest and matched against previously recorded fingerprint to identify the repeated offender. Forensic experts also found fingerprints very helpful to identify the criminals by matching the latent fingerprint left at the scene of crime with previously recorded fingerprints. The government law enforcement agencies extensively sponsored the scientific study and development of visual matching techniques for fingerprint matching. They also sponsored training programs for fingerprint experts. The art of fingerprint matching technique was successfully applied to identify the law breakers \cite{scott1951fingerprint,gaensslen2001advances}.
	
Extensive research and training has been accomplished to increase the efficiency and precision of manual fingerprint indexing and matching technique. Due to robustness to forgery, there was a very high demand for fingerprint experts, but the limited number of experts available were not able to cope with the growing demands. The fingerprint indexing techniques based on the \textit{Henry system} of fingerprint classification generate an extremely skewed distribution of fingerprint in five sets \cite{maltoni2009handbook}. The training of fingerprint experts was very time consuming and tedious, this made the availability of fingerprint experts very limited, but the demand was very high. All this promoted the research in the area of electronically acquisition of fingerprint data and fully automated fingerprint recognition systems. The dedicated effort of researchers leads to development of \textit{Automatic Fingerprint Identification System (AFIS)} over the last $5$ decades. The low priced and easy availability of highly sophisticated and powerful handheld electronic devices with advanced wireless communication facilities has increased the public concern about information security and identity fraud, resulting in an increase in the demand of fingerprint biometric and other biometric based access control mechanisms for non-forensic applications.

\section{Survey Methodology}

This survey presents a comprehensive analysis of fingerprint orientation field (OF) estimation techniques across classical, statistical, and deep learning–based paradigms. The methodology adopted to identify, classify, and evaluate these techniques is described below.

\subsection{Literature Search and Data Sources}

A systematic literature review was conducted using prominent academic databases and digital libraries, including \textit{IEEE Xplore}, \textit{ACM Digital Library}, \textit{ScienceDirect}, \textit{SpringerLink}, and \textit{Google Scholar}. The search queries included combinations of keywords such as \textit{``fingerprint orientation estimation''}, \textit{``ridge flow modeling''}, \textit{``orientation field estimation''}, \textit{``latent fingerprint enhancement''}, and \textit{``deep learning for fingerprints''}.

Relevant publications were retrieved using Boolean filters and citation tracking, focusing on papers published between 1993 and 2025. Preference was given to peer-reviewed journal articles, high-impact conference proceedings, and highly cited preprints in the biometric domain.

\subsection{Inclusion and Exclusion Criteria}

To ensure rigor and relevance, the following inclusion criteria were applied:
\begin{itemize}
    \item Studies proposing novel methods or significant improvements in fingerprint orientation field estimation.
    \item Articles with clearly described mathematical models and empirical validation.
    \item Papers applying classical, statistical, or machine learning techniques for orientation estimation.
\end{itemize}

Studies were excluded if:
\begin{itemize}
    \item The focus was solely on fingerprint matching or enhancement without addressing of estimation.
    \item Empirical results or algorithmic details were insufficient or unverifiable.
    \item The work was not peer-reviewed or lacked scholarly attribution.
\end{itemize}

\subsection{Temporal and Thematic Scope}

This review encompasses research published over a span of more than three decades, starting with foundational gradient-based methods in the early 1990s~\cite{sherlock1993model} to state-of-the-art deep learning frameworks in the mid-2020s~\cite{gao2018orienet, tang2017fingernet}. The survey aims to capture the progression of estimation techniques in response to emerging challenges such as noise resilience, partial fingerprint coverage, and computational scalability.

\subsection{Classification Framework}

To facilitate structured analysis, the collected literature was categorized into the following classes:
\begin{enumerate}
    \item \textbf{Gradient-based Methods:} Traditional approaches based on image derivatives and directional filtering~\cite{rao1990identification}.
    \item \textbf{Model-based Methods:} Techniques using polynomial fitting, Legendre, Chebyshev, and Fourier expansions~\cite{wang2007fingerprint, ram2009active}.
    \item \textbf{Probabilistic Models:} Estimations derived using Markov Random Fields (MRF) and statistical constraints~\cite{lee2008orientation}.
    \item \textbf{Dictionary Learning Techniques:} Patch-based orientation synthesis methods using similarity constraints~\cite{feng2012orientation}.
    \item \textbf{Neural Network and Deep Learning Approaches:} CNNs, DRNNs, and hybrid architectures trained for classification or regression~\cite{cao2015latent, schuch2017convnet}.
    \item \textbf{Hybrid and Multi-Scale Methods:} Combinations of classical and learning-based methods that integrate spatial coherence and multi-resolution processing~\cite{chen2013multi}.
\end{enumerate}

Each category was analyzed based on:
\begin{itemize}
    \item Mathematical formulation and computational complexity.
    \item Robustness to image quality degradation and presence of noise.
    \item Applicability to partial, latent, or overlapped fingerprints.
    \item Usage of standard fingerprint databases such as NIST SD27, FVC2004/2006, and CASIA-A~\cite{fvc2004, casia2005}.
\end{itemize}

\subsection{Performance Evaluation Criteria}

Where applicable, evaluation metrics such as root mean square error (RMSE), Equal Error Rate (EER), False Match Rate (FMR), False Non-Match Rate (FNMR), and singularity point detection accuracy (Precision, Recall, and F-measure) were extracted. These metrics allow comparative insights into the effectiveness of different OF estimation approaches.

\subsection{Synthesis and Discussion Strategy}

Based on the classified and evaluated methods, a comparative discussion is presented in subsequent sections. Key aspects analyzed include:
\begin{itemize}
    \item Evolutionary trends and methodological shifts from handcrafted to learning-based approaches.
    \item Trade-offs between accuracy, generalization, and computational cost.
    \item Practical deployment challenges in large-scale and embedded biometric systems.
\end{itemize}

Through this structured methodology, we aim to provide a holistic and insightful synthesis of existing fingerprint orientation estimation techniques, identify unresolved challenges, and highlight promising directions for future research.

\section{Orientation Estimation of Fingerprint}\label{or}
		
For successfully implementation of the fingerprint biometric system the orientation estimation of fingerprint ridge is a crucial step. Its primary aim is to determine the ridge flow direction in localized areas of a fingerprint image, which significantly aids in feature extraction, noise reduction, and overall enhancement of matching accuracy. This step is essential for preprocessing tasks, including segmentation, thinning, and minutiae detection.
	
\begin{figure}[!h]
			\centering

			\includegraphics[scale=0.5]{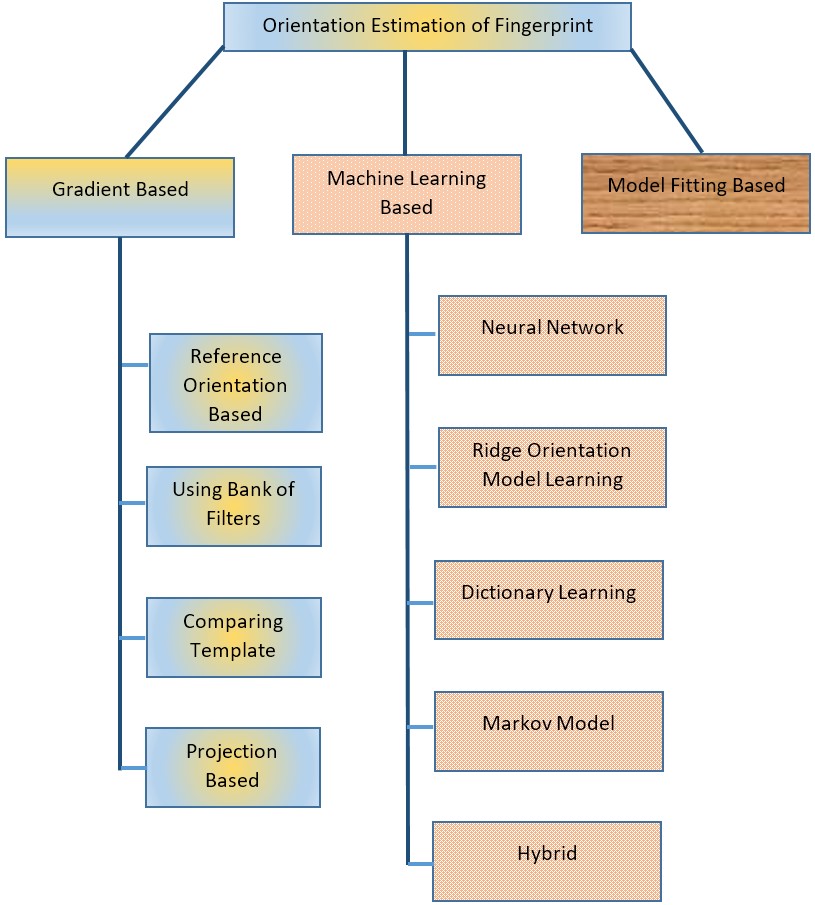}

			\caption{Classification of Fingerprint Orientation Techniques}
			\label{fig:class}
		\end{figure}

	The fingerprint orientation estimation techniques presented in literature can be classified into three main categories namely: \textit{\textbf{gradient based, machine learning and model fitting based}} as shown in the figure \ref{fig:class}. The gradient-based infer ridge orientations using mathematical operation on intensity of fingerprint image pixxels. These approaches analyze the local intensity gradients in fingerprint images to identify the dominant direction of ridge flow. The simplicity and computational efficiency of gradient-based methods make them a popular choice for real-time applications. However, their performance can degrade in low-quality fingerprint images where noise and distortions obscure ridge details.
	
	Advanced techniques such as tensor analysis, machine learning techniques, frequency based  complex filters address the problems gradient based technique for low-quality image.  Tensor-based methods calculate the orientation field by considering the coherence of ridge patterns across larger regions, resulting in smoother and more consistent estimates. Frequency-domain techniques use Fourier analysis to capture periodic ridge structures, effectively handling degraded areas of the image.
	
	Emerging techniques in field of artificial intelligence incorporate machine learning and deep learning for orientation estimation, leveraging large datasets to train models capable of predicting ridge flows with high accuracy. These approaches are particularly adept at handling complex noise, smudges, and distortions in fingerprints. By combining spatial and contextual information, deep learning-based models generate robust and globally consistent orientation fields, making them suitable for modern applications like touchless fingerprint scanning.

	A Fingerprint is composed of ridge lines, which look like curve lines if the width is ignored. The orientation of these curves can be used to remove the irregularity due to noise. The orientation information of the ridges can also be used as a feature. So the efficient and accurate estimation of the orientation of a fingerprint will be very useful.
	
\begin{figure}[!ht]
		\centering
		$\begin{array}{ccc}
		\includegraphics[scale=0.61]{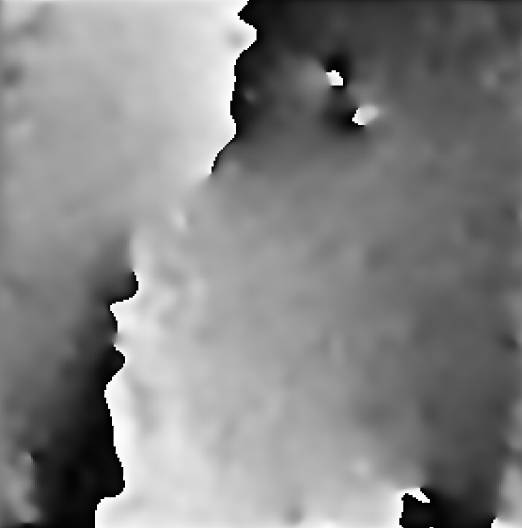}&&
		\includegraphics[scale=0.45]{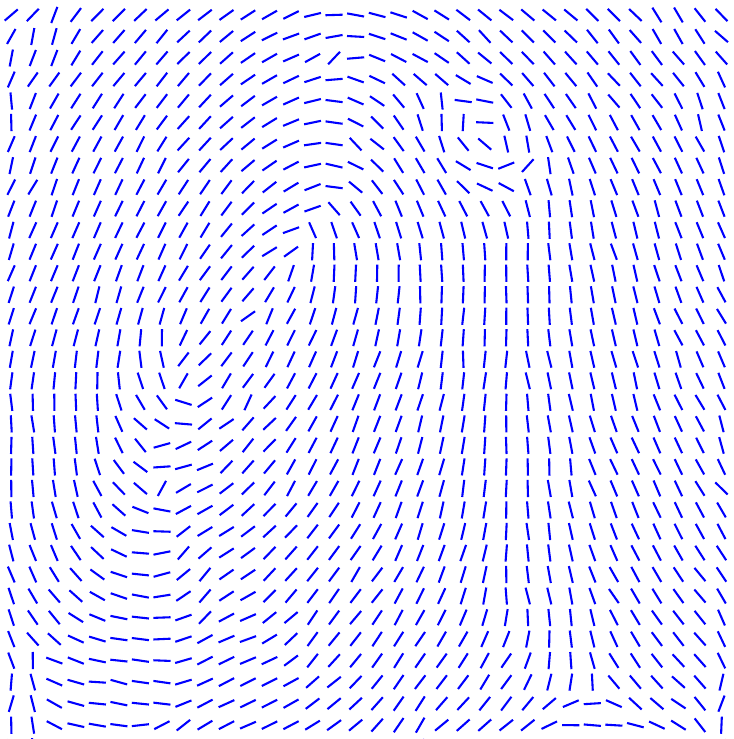}\\
		(a)&& (b) \\
		\end{array}$
		\caption{Orientation image: a) Image with a value of gradient as pixel intensity and b) average orientation into 10 x 10 mesh grid }
		\label{fig:or_img}
	\end{figure}

	The angle $\theta_{ij}$, between the small portion of the ridge line centered at $(i, j)$ and horizontal axis is called orientation of the ridge at $(i, j)$. The ridge line does not have an associated start and end point, so it is undirected and values of $\theta_{ij}$ lies in interval $[\theta ~\pi[$. The orientation of the ridge hardly changes at each pixel location, so it is useless to do the computationally expansive work of calculation of orientation for each pixel. Instead, the fingerprint is divided into small square-meshed grid and orientation is estimated for a central pixel of each cell. The matrix composed of these estimated orientations is called \textit{orientation image} or \textit{directional image} or \textit{orientation map} \cite{grasselli1969automatic}, see Figure \ref{fig:or_img}. 

\begin{landscape}

\begin{longtable}{p{4.5cm} p{5.5cm} p{5.5cm} p{1.5cm}}
\caption{Orientation Estimation of Fingerprint} \label{tab:orientation_survey1} \\
\toprule
\textbf{Category} & \textbf{Sub-Category} & \textbf{Author(s)} & \textbf{Year} \\
\midrule
\endfirsthead

\caption[]{Orientation Estimation of Fingerprint (continued)} \\
\toprule
\textbf{Category} & \textbf{Sub-Category} & \textbf{Author(s)} & \textbf{Year} \\
\midrule
\endhead

\midrule \multicolumn{4}{r}{\textit{Continued on next page}} \\
\midrule
\endfoot

\bottomrule
\endlastfoot
	 \multirow{3}{3cm}{\textbf{Gradient Based Techniques}}&\multirow{27}{*}{\textbf{}}
	 &Grasselli \cite{grasselli1969automatic} &1969\\

	 \multirow{25}{*}{}
	 &&Stock \cite{stock1969development}&1969\\
	
	 &&Kars \cite{kass1987analyzing} & 1987\\
	
	 &&Mehtre \cite{mehtre1987segmentation}&1987\\
	
	 &&Rao \cite{rao2012taxonomy} & 1990 \\

	 &&Donahue \cite{donahue1993use}&1993\\

	 &&Rath \cite{ratha1995adaptive}&1995\\

	 &&Kamei \cite{kamei1995image}&1995\\

	 &&Da Costa \cite{da2001new}&2001\\

	 &&Bazen \cite{bazen2002systematic}&2002\\

	 &&He \cite{he2003image}&2003\\
	
	 &&Kamei \cite{kamei2004image}&2004\\

	 &&Nakamura \cite{nakamura2004fingerprint}&2004\\
	
	 &&Shertock \cite{sherlock2004computer}&2004\\

	 &&Jian et al.[50]&2005\\
	
	 &&Chikkerur \cite{chikkerur2007fingerprint}&2007\\
	
	 &&Wang \cite{wang2007enhanced}&2007\\
	
	 &&Kekre and Bharadi\cite{kekre2009fingerprint}&2009\\
	
	 &&Oliveira \cite{oliveira2008multiscale}&2008\\
	 
	 &&Ji \cite{ji2008fingerprint}&2008\\
	
	 &&Mei et al.\cite{mei2009gradient}&2009\\
	
	 &&Babatunde ey al. \cite{babatunde2012block}&2012\\
	
	 &&Bian et al.\cite{bian2014fingerprint}&2014\\
	 
	 &&Sulong et al.\cite{sulong2016technique}&2016\\
	
	 &&Bian et al.\cite{bian2017combining}&2017\\
	
	 &&Li et al. \cite{li2018combining}&2018\\
	
	 &&Abbood et al. \cite{abbood2018new}&2018\\

	 &&AlShehri et al.\cite{alshehri2019alignment}&2019\\
	
	 &&Duan et al. \cite{duan2021orientation}&2021\\
	
	 &&Shen et al. \cite{shen2023effective}&2023\\
    
     &&Cappelli et al. \cite{cappelli2024exploring}&2024\\
	 
	 &&Prakash et al. \cite{prakash2024unbalanced}&2024\\

	 \multirow{2}{*}{\textbf{}}&\multirow{2}{3cm}{\textbf{Reference Orientation Based}}
	 &Oliveira \cite{oliveira2008multiscale} & 2008\\
	
	 &&Ji \cite{ji2008fingerprint}&2008\\

	 \multirow{2}{*}{\textbf{}}&\multirow{2}{3cm}{\textbf{Using banks of Filters banks}}
	 & Hong \cite{hong96}& 1996 \\
	
	 && & \\

	 \multirow{2}{*}{\textbf{}}&\multirow{2}{3cm}{\textbf{Comparing Templates}}
	 &Shmurun et al. \cite{shmurun1994extraction} &1994 \\
	
	 &&Kawagoe \cite{kawagoe1984fingerprint}&1984\\
	
	 \multirow{3}{*}{\textbf{}}&\multirow{1}{3cm}{\textbf{Projection based}}
	 & Shrlock \cite{shrlock1994}&1994 \\

	 \multirow{3}{3cm}{\textbf{Machine Learning Based Techniques}}&\multirow{2}{3cm}{\textbf{Neural Network}}
	 &Zhu et al.\cite{zhu2006systematic}&2006\\
	
	 &&Ji et al.\cite{ji2008fingerprint}&2008\\
	 
	 &&Sahasrabudhe \cite{sahasrabudhe2013learning}&2013\\
	 
	 &&Wu et al. \cite{wu2013svm}&2013\\

	 &&Cao et al.\cite{cao2015latent}&2015\\

	 &&Schuch et al.\cite{schuch2017convnet}&2017\\
	
	 &&Tang et al. \cite{tang2017fingernet}&2017\\
	
	 &&Qu et al. \cite{qu2018orienet}&2018\\

	 \multirow{2}{*}{\textbf{}}&\multirow{2}{3cm}{\textbf{Ridge orientation model learning}}
	 &Ram et al. \cite{ram2009active}&2009\\
	
	 && Zhang et al.\cite{zhang2014overlapped}&2014\\
	
	 && Zhang et al.\cite{zhang2014adaptive}&2014\\

	 \multirow{2}{*}{\textbf{}}&\multirow{2}{3cm}{\textbf{Dictionary learning}}
	 &Feng et al.\cite{feng2012orientation}&2012\\
	 
	 &&Yang et al.\cite{yang2014localized}&2014\\
	
	 &&Jain et al.\cite{jain2015fingerprint}&2015\\
	
	 &&Chen et al.\cite{chen2016multi}&2016\\
	
	 &&Cao et al. \cite{cao2014segmentation}&2014\\
	
	 &&Liu et al \cite{liu2017sparse}&2017\\

	 \multirow{2}{*}{\textbf{}}&\multirow{1}{3cm}{\textbf{Markov Model}}
	 &Lee et al.\cite{lee2008probabilistic}&2008\\
	
	 \multirow{2}{*}{\textbf{}}&\multirow{1}{3cm}{\textbf{Hybrid}}
	 &Turroni et al.\cite{turroni2011improving}&2011\\

	 \multirow{3}{3cm}{\textbf{Model Fitting Based}}&\multirow{1}{*}{\textbf{}}
	 &Sherlock et al.\cite{sherlock1993model}&1993\\
	 
	 &&Vizcaya et al.\cite{vizcaya1996nonlinear}&1996\\
	 
	 &&Zhou et al. \cite{zhou2004modeling}&2004\\
	
	 &&Gu et al.\cite{gu2004combination}&2004\\
	 
	 & &Zhou et al.\cite{zhou2004model}&2004\\
	
	 & &Li et al.\cite{li2006constrained}&2006\\
	
	 & &Li et al.\cite{li2007stability}&2007\\
	 
	 &&Huckemann et al.\cite{huckemann2008global}&2008\\
	
	 & &Gottschlich et al.\cite{gottschlich2016perfect}&2016\\
	
	 & &Wang et al.\cite{wang2007fingerprint}&2007\\
	
	 & &Wang et al. \cite{wang2010global}&2010\\
	 
	 & &Tashk et al.\cite{tashk2009improvement}&2009\\
	
	 & &Tao et al.\cite{tao2010estimation}&2010\\
	
	 & &Ram et al.\cite{ram2008curvature}&2008\\
	
	 & &Ram et al.\cite{ram2010modelling}&2010\\
	
	 & &Tashk et al.\cite{tashk2010chebyshev}&2010\\
	
	 & &Jirachaweng et al.\cite{jirachaweng2011residual_1}&2011\\
	
	 & &Lui and Lui\cite{liu2012fingerprint}&2012\\
	
	 & &Liu et al.\cite{liu2014fingerprint}&2014\\

	 & &Gupta et al.\cite{gupta2015fingerprint}&2015\\
	
	 & &Gupta et al.\cite{gupta2016accurate}&2016\\
	
	 & &Bian et al.\cite{bian2014fingerprint}&2014\\

	\end{longtable}
    
\end{landscape}

\begin{landscape}
 
\begin{longtable}{ p{3.5cm}p{3cm}p{4cm}p{2.5cm}p{1.2cm} }
\caption{Orientation Estimation of Fingerprint Methods} \label{tab:orientation_survey} \\
\hline
\textbf{Technique Type} & \textbf{Sub-Category} & \textbf{Author(s)} & \textbf{Performance / Dataset} & \textbf{Year} \\
\hline
\endfirsthead

\multicolumn{5}{c}%
{{\bfseries Table \thetable\ Continued from previous page}} \\
\hline
\textbf{Technique Type} & \textbf{Sub-Category} & \textbf{Author(s)} & \textbf{Performance / Dataset} & \textbf{Year} \\
\hline
\endhead

\hline \multicolumn{5}{|r|}{{Continued on next page}} \\ \hline
\endfoot

\hline
\endlastfoot

\multirow{11}{3cm}{Gradient Based} & Classic Gradient Method & Grasselli \cite{grasselli1969automatic} & N/A & 1969 \\

& Classic Gradient Method & Stock \cite{stock1969development} & N/A & 1969 \\

 & Classic Gradient Method & Kass \cite{kass1987analyzing} & N/A & 1987 \\

 & Classic Gradient Method & Mehtre \cite{mehtre1987segmentation} & N/A & 1987 \\

& Classic Gradient Method & Rao \cite{rao2012taxonomy} & N/A & 1990 \\

 & Classic Gradient Method & Donahue \cite{donahue1993use} & N/A & 1993 \\

& Classic Gradient Method & Ratha \cite{ratha1995adaptive} & N/A & 1995 \\

 & Classic Gradient Method & Bazen \cite{bazen2002systematic} & N/A & 2002 \\

 & Classic Gradient Method & Oliveira \cite{oliveira2008multiscale} & FVC2002 & 2008 \\

 & Classic Gradient Method & Mei et al. \cite{mei2009gradient} & N/A & 2009 \\

& Classic Gradient Method & Li et al. \cite{li2018combining} & N/A & 2018 \\

\multirow{7}{3cm}{Model Fitting} & Orientation Field Modeling & Sherlock et al. \cite{sherlock1993model} & N/A & 1993 \\

 & Orientation Field Modeling & Zhou et al. \cite{zhou2004modeling} & FVC2004 & 2004 \\

& Orientation Field Modeling & Ram et al. \cite{ram2008curvature} & NIST SD27 & 2008 \\

 & Orientation Field Modeling & Gupta et al. \cite{gupta2015fingerprint} & FVC2002 & 2015 \\

& Polynomial Field Estimation & Gottschlich \cite{gottschlich2016perfect} & FVC2004 & 2016 \\

 & Chebyshev Polynomial & Tashk et al. \cite{tashk2010chebyshev} & NIST SD27 & 2010 \\

 & Residual Minimization & Jirachaweng et al. \cite{jirachaweng2011residual_1} & FVC2002 & 2011 \\

\multirow{7}{3cm}{Machine Learning} & Deep Learning (CNN) & Tang et al. \cite{tang2017fingernet} & FVC2004, Accuracy: 94.2\% & 2017 \\

 & Deep Learning (CNN) & Qu et al. \cite{qu2018orienet} & NIST SD27, Accuracy: 93.8\% & 2018 \\

 & Deep Learning (CNN) & Shen et al. \cite{shen2023effective} & FVC2006, Accuracy: 95.1\% & 2023 \\

 & Deep Learning (CNN) & Prakash et al. \cite{prakash2024unbalanced} & FVC2002, Cross-Sensor Study & 2024 \\

 & Dictionary Learning & Feng et al. \cite{feng2012orientation} & FVC2002 & 2012 \\

 & Ridge Orientation Model & Zhang et al. \cite{zhang2014adaptive} & FVC2004 & 2014 \\

 & Hybrid Neural-SVM & Wu et al. \cite{wu2013svm} & NIST SD27 & 2013 \\

Filter-Based & Gabor Filter Bank & Hong et al. \cite{hong96} & FVC2000 & 1996 \\

Reference-Based & Template Matching & Kawagoe \cite{kawagoe1984fingerprint} & N/A & 1984 \\

Projection-Based & Orientation via Projections & Sherlock \cite{shrlock1994} & N/A & 1994 \\

\end{longtable}
   
\end{landscape}
The table titled \textit{Orientation Estimation of Fingerprint Methods} provides a comprehensive overview of various approaches developed over time for estimating fingerprint orientation fields—an essential preprocessing step in fingerprint recognition systems. This tabulated survey categorizes methods by their underlying technique types, sub-categories, authorship, dataset utilization, performance metrics (when available), and year of publication. The methods are grouped into five main technique types: Gradient Based, Model Fitting, Machine Learning, Filter-Based, Reference-Based, and Projection-Based, with special emphasis on recent deep learning approaches.

\textbf{Gradient-based methods} are among the earliest approaches and remain widely referenced due to their simplicity and computational efficiency. These techniques rely on local image gradients to estimate ridge orientations. Notable early contributions include Grasselli \cite{grasselli1969automatic} and Mehtre \cite{mehtre1987segmentation}, which laid the groundwork for directional estimation using local derivatives. Over the decades, these methods have evolved with more refined gradient computation techniques (e.g., Mei et al. \cite{mei2009gradient}), often aiming for increased robustness in low-quality images.

\textbf{Model fitting methods} represent another classical category, where orientation fields are modeled globally using polynomial or parametric forms. Sherlock et al. \cite{sherlock1993model} and Zhou et al. \cite{zhou2004modeling} introduced systematic ways to impose global consistency on orientation fields through mathematical modeling. These methods typically demonstrate higher resilience to local image noise and irregularities. For example, Gottschlich \cite{gottschlich2016perfect} uses polynomial field estimation to generate ``perfect'' orientation fields, improving the continuity and realism of orientation maps in low-quality images.

\textbf{Machine learning methods}, and particularly \textbf{deep learning-based techniques}, are more recent and represent a significant advancement in the field. Highlighted with a gray background in the table, these methods utilize convolutional neural networks (CNNs) to learn complex features and orientation patterns directly from data. Tang et al.'s \textit{FingerNet} \cite{tang2017fingernet} and Qu et al.'s \textit{OrienNet} \cite{qu2018orienet} achieved accuracy levels exceeding $93\%$ on benchmark datasets such as FVC2004 and NIST SD27. These methods offer state-of-the-art performance and are highly scalable, benefiting from modern GPU computing. More recent contributions like Shen et al. \cite{shen2023effective} and Prakash et al. \cite{prakash2024unbalanced} further optimize orientation estimation through balanced training and cross-sensor generalization.

Other sub-categories within machine learning include dictionary learning (Feng et al. \cite{feng2012orientation}) and hybrid methods combining neural networks with support vector machines (Wu et al. \cite{wu2013svm}). These approaches often target specific limitations of traditional methods, such as poor generalization in noisy or distorted images.

The table also includes \textbf{filter-based} methods such as the Gabor filter bank approach by Hong et al. \cite{hong96}, which combines directional filtering with frequency analysis to enhance and estimate ridge flow. Additionally, \textbf{reference-based} methods like template matching (Kawagoe \cite{kawagoe1984fingerprint}) and \textbf{projection-based} methods (Sherlock \cite{shrlock1994}) offer alternative strategies, though they are less commonly used in modern systems due to their limited adaptability.

In summary, this table not only captures the chronological evolution of fingerprint orientation estimation but also highlights methodological shifts—from handcrafted mathematical models to data-driven learning-based methods. The inclusion of performance metrics and dataset references provides valuable insights into the applicability and generalizability of each method, making the table a useful reference for both researchers and practitioners in biometric recognition.

\subsection{Gradient Based Techniques}

The gradient-based approach is a method used to compute the orientation or angle of ridge lines in a fingerprint image. This is crucial in fingerprint biometric system. The goal is to identify the direction of an ridge line at each pixel.

Mathematically the gradient at pixel $[x, y ]$  can be represented as(Eq \ref{eq:grad}) :
\begin{equation}
\theta(x,y) = \frac{1}{2} \tan^{-1}\left(\frac{2 \cdot G_{xy}(x,y)}{G_{xx}(x,y) - G_{yy}(x,y)}\right)
\label{eq:grad}
\end{equation}

Here, \( G_{xx} \), \( G_{yy} \), and \( G_{xy} \) are the elements of the gradient tensor, representing second-order partial derivatives of the image. Specifically: \( G_{xx} = (I_x)^2 \) represents the square of the image gradient along the \(x\)-axis, indicating changes in the horizontal direction and \( G_{yy} = (I_y)^2 \) represents the square of the gradient along the \(y\)-axis, showing variations in the vertical direction. \( G_{xy} = I_x \times I_y \) represents the product of gradients along both axes, capturing the relationship between the horizontal and vertical changes.

The term \( \frac{2 \cdot G_{xy}}{G_{xx} - G_{yy}} \) in the formula captures the strength and direction of edges. The function \( \tan^{-1} \) (inverse tangent) converts this ratio into an angle, giving the orientation of the edge at each point. Finally, dividing by 2 accounts for the symmetrical nature of the gradient. This method is widely used in edge detection algorithms, like the Canny edge detector, to determine the direction of edges for further processing.

The ridge orientation can be calculated from the gradient values at each pixel value in fingerprint image. The partial-derivatives, $[\nabla_{i}(i,j), \nabla_{j}(i,j)]$ represent the gradient at the pixel $[i, j]$ of fingerprint image $I$. $\nabla_{i}$ and $\nabla_{j}$ are partial derivatives of $I$ with respect to $i$ and $j$ in respective directions at pixel $[i, j]$. The orientation $\theta_{ij}$ is calculated as the \textit{arctangent} of $\nabla_{i}/\nabla_{j}$. This method is very simple, but have two disadvantages: i) discontinuity and non-linearity at $\pi/2$ and ii) circularity of angle make averaging meaningless.
\subsubsection{Weighted Gradient Orientation Estimation}
In Weighted Gradient Orientation Estimation, each gradient vector is weighted based on local intensity or coherence, allowing areas with strong ridge structures to contribute more to the orientation estimate \cite{1587631}. If $w_i$
is the weight assigned to each pixel 
$i$ in a block, the weighted gradients in direction of $x$, $\mathcal{G}_{x,weighted}$  and in direction of $y$, $\mathcal{G}_{y,weighted}$ are calculated using the Eq. \ref{eq:gx} and Eq. \ref{eq:gy}.
\begin{equation}
\mathcal{G}_{x,weighted}=\frac{\Sigma_{i=1}^{N}w_i\mathcal{G}_{x,i} }{\Sigma_{i=1}^{N}w_i}
\label{eq:gx}
\end{equation}
\begin{equation}
\mathcal{G}_{y,weighted}=\frac{\Sigma_{i=1}^{N}w_i\mathcal{G}_{y,i} }{\Sigma_{i=1}^{N}w_i}
\label{eq:gy}
\end{equation}
Based on the weighted gradients of pixel $i$, the orientation for the block is estimated using Eq. \ref{eq:bo}.
\begin{equation}
\theta_{block}=arctan(\frac{\mathcal{G}_{y,weighted}}{\mathcal{G}_{x,weighted}})
\label{eq:bo}
\end{equation} 
\begin{table}[!ht]
\centering
\caption{Classification of gradient based techniques}
\begin{tabular}{|p{4cm}|p{2.5cm}|p{2cm}|p{2cm}|p{4cm}|}
\hline
\textbf{Technique}&\textbf{Computational Efficiency}&\textbf{Accuracy}&\textbf{Robustness}&\textbf{Best Use Case}\\
\hline
Weighted Gradient Orientation Estimation&High&High&Moderate&Variable quality fingerprints with inconsistent ridge flow\\
\hline
Averaged Gradient Field (AGF)&Moderate&Low&High&Real-time processing in uniform, high-quality images\\
\hline
Gaussian Gradient-Based Orientation Estimation&High&High&Moderate&Noisy or low-quality fingerprints\\
\hline
Gradient Coherence Orientation Estimation&High&Moderate to High&Moderate&High-coherence regions with consistent ridges\\
\hline
Gradient-Based Tensor Analysis&High&High&Moderate&Complex or noisy fingerprints\\
\hline
Gradient Vector Field Analysis&Moderate&Low&High&High-quality fingerprints with smooth ridge flow\\
\hline
Reference Orientation-Based&High&Moderate&Moderate&High-quality images in controlled environments\\
\hline
Using Banks of Filters&Very High&Very High&Moderate to Low&Noisy/low-quality images with sufficient processing power\\
\hline
Template Comparison-Based&High&High&Moderate&Complex/noisy fingerprints; sufficient resources available\\
\hline
Projection-Based Technique&Moderate to High&Moderate&High&Real-time processing with clear, high-quality images\\
\hline
\end{tabular}
\end{table}

	For solving the above problem of circularity of angles, Kass et al. suggested to double the angles \cite{kass1987analyzing}. Based on this idea many efficient methods are proposed for computing the \textit{orientation image of fingerprint} \cite{rao2012taxonomy, ratha1995adaptive, bazen2002systematic}. Donhue et al. also proposed a method based on gradient independently in $1993$ \cite{donahue1993use}. The perform the least-square minimization for finding the average orientation. 
	
	The gradient based orientation estimation fails in a high noise sensitive area where the gradient tends to zero. To solve this problem second order derivatives were suggested \cite{larkin2005uniform}. But it solves the problem partially. Da Costa et al. select first or second order derivatives based on the local coherence of the two \cite{da2001new}. 
	
	The optimum output of different many directional filters in frequency domain ($16$ numbers of direction filters) were used by \cite{kamei2004image, kamei1995image}. Hong et al. \cite{hong96} and Nakamura et al. \cite{nakamura2004fingerprint} proposed method for spatial domain based on Gabor filter that generate analogous result. 
Based on \textit{Short Time Fourier Transform} (STFT) was proposed by Chikkerur et al. \cite{chikkerur2007fingerprint}.
The Short-Time Fourier Transform (STFT) is a technique used to analyze signals in both the time (or spatial) domain and frequency domain. In the context of fingerprint enhancement, as discussed in the paper by Chikkerur et al. \cite{chikkerur2007fingerprint}, the STFT is employed to estimate the local ridge frequency of a fingerprint image. The ridge frequency is a crucial feature for enhancing the image and improving fingerprint recognition accuracy.

The STFT is given by (Eq. \ref{eq:sift}):
\begin{equation}
STFT(u,v) = \sum_{x,y} f(x,y) w(x-u,y-v) e^{-j2\pi (ux+vy)}
\label{eq:sift}
\end{equation}

Here, \( f(x,y) \) represents the image or signal, \( w(x,y) \) is a window function, and \( e^{-j2\pi (ux+vy)} \) is the complex exponential that defines the frequency components. The variables \( u \) and \( v \) correspond to the frequency components in the horizontal and vertical directions, respectively.

The window function \( w(x-u, y-v) \) helps localize the Fourier transform to a small region around a point \((u,v)\). It controls the trade-off between time and frequency resolution, allowing the analysis of local features in the image, which is particularly useful in fingerprint processing where ridge patterns can vary across the image. By applying the STFT, the authors are able to estimate the local ridge frequency, which corresponds to the periodicity of the ridges in the fingerprint. This information is used for enhancing the fingerprint image, making ridge structures more prominent and aiding in feature extraction for better matching in fingerprint recognition systems. This approach significantly improves the accuracy of fingerprint enhancement, especially in cases with noisy or degraded fingerprint images.

 Wang et al. \cite{wang2007enhanced} proposed an implementation of gradient-based technique for fingerprint orientation estimation to improve the performance. For each fingerprint image block, the best orientation is chosen from the four overlapping blocks based on the voting of reliability measures. Kekre and Bharadi \cite{kekre2009fingerprint} proposed a gradient based technique and for smoothing the orientation they perform neighborhood averaging. 
	Ali Ismail Awad has presented an implementation of gradient based technique using Graphics Processing Unit (GPU) \cite{awad2016fast} , which was $6.41$ times faster than CPU based implementation as claimed by \cite{awad2016fast}. Dyre and Sumathi presented a gradient based orientation estimation smoothing technique based on analysis of the consistency of orientation in the neighborhood. Abbod et al. \cite{abbood2018new} divide the fingerprint image into a block of size $16 \times 16$. Before calculation of the gradient, they applied a Epicycloid shape filter on the target block to mask some pixel.

	\subsubsection{Reference Orientation Based Techniques}
	A very simple technique is using $n$ numbers of reference orientation \cite{maltoni2009handbook}.
	
	\begin{equation}
	\theta_{k}=k\dfrac{\pi}{n} 
	\end{equation}
	
	where $k=0... n-1$, and $n$ is the number of reference orientation. Based on the pixel intensity along the reference orientation $\theta_{k_{opt}}$ is selected. The local orientation $\theta_{ij}=\theta_{k_{opt}}$ at the pixel $I(i, j)$ is computed on a window $\mathcal{W}$ centered at $I(i, j)$. The sum of pixel intensity along the reference orientation in window $\mathcal{W}$ is minimum and maximum for ridge and valley in the direction of the ridge and valley respectively \cite{stock1969development, mehtre1987segmentation, he2003image, oliveira2008multiscale}. The \textit{standard deviation} of the intensity of the pixels along the reference orientation is calculated and the optimal reference orientation is selected depending on the maximum standard deviation contrast between a reference orientation and its orthogonal orientation \cite{oliveira2008multiscale}. It was suggested by Sherlock \cite{sherlock2004computer} to optimized the projection of the ridge line inside the window $\mathcal{W}$ along a number of reference orientation. The reference orientation corresponding to the smallest variation in the projection is taken as orientation of the ridge. Ji et al. modify the above approach by removing the central ridge line before computing the projections \cite{ji2008fingerprint}. 
	
	Both gradient based technique and reference orientation based techniques have some advantage and disadvantage. The computational complexity of gradient based technique is less than reference orientation based techniques. In reference based technique, a probability value can be assigned to a different orientation value that may help in the next level of processing to smooth the estimated orientation image.
	
\subsubsection{Comparing Templates}
Shmurun et al. \cite{shmurun1994extraction} presented a template matching technique for construction of the ridge orientation map from a gray image of the fingerprint. They used the radial basis function to reduce the number of base templates. The orientation map was calculated from the contribution of base templates.

\subsection{Machine Learning Based Techniques}
\subsubsection{Neural Network}

Nagaty \cite{nagaty2003learning} presented a hierarchical neural network of a \textit{Back Propagation Network} (BPN) and \textit{Self-organized Feature Maps} (SOFM). BPN was used for feature extraction from blocks of the image and SOFM was used for clustering. The BPN learns weights \( \mathbf{W} \) by minimizing the loss function:

\[
L = \frac{1}{N} \sum_{i=1}^{N} \| \hat{y}_i - y_i \|^2,
\]

where \( \hat{y}_i \) is the network output and \( y_i \) is the target output. SOFM was used to cluster feature vectors \(\mathbf{x}\) by updating weights \(\mathbf{w}_j\) using:

\[
\Delta \mathbf{w}_j = \eta \, h_{j,i} \, (\mathbf{x}_i - \mathbf{w}_j),
\]

where \( h_{j,i} \) is the neighborhood function. The system was robust for noisy fingerprint image. 

Zhu et al. \cite{zhu2006systematic} applied the neural network for the evaluation of correctness of orientation estimated by gradient-based method. The local ridge orientation \(\theta\) was estimated using gradients:

\[
\theta = \frac{1}{2} \arctan\left( \frac{2S_{xy}}{S_{xx} - S_{yy}} \right),
\]

where \(S_{xx}, S_{yy}, S_{xy}\) are local gradient sums. The neural network was trained to differentiate between correct and incorrect ridge orientation estimation. The correct orientation of the block around the incorrect orientation block is used for correction of the error. 

Ji et al. \cite{ji2008fingerprint} presented a method for calculation of four-direction fingerprint orientation field. They applied \textit{Pulse Coupled Neural Network} (PCNN) for identification of the primary ridge in a block of fingerprint images. PCNN neuron dynamics were described by:

\[
F_{ij}(t) = S_{ij} + \sum_{(k,l) \in N_{ij}} W_{kl} Y_{kl}(t-1),
\]

\[
Y_{ij}(t) = 
\begin{cases}
1, & \text{if } F_{ij}(t) > \theta_{ij}(t) \\
0, & \text{otherwise}
\end{cases},
\]

where \(F_{ij}\) is the internal activity and \(\theta_{ij}\) is the dynamic threshold. They claimed that the proposed technique is able to achieve the good accuracy on the FCV2004 database with low computational time. 

For the restoration of noisy fingerprint with the help of orientation map, Sahasrabudhe et al. proposed an orientation estimation technique \cite{sahasrabudhe2013learning}. They applied two \textit{Continuous Restricted Boltzmann Machines} (CRBM) to learn the orientation pattern of a fingerprint. The CRBM models the joint probability as:

\[
E(v, h) = \frac{1}{2\sigma^2} \| v - W h - b \|^2 - c^\top h,
\]

where \(v\) is the visible unit (orientation patch), \(h\) the hidden unit, and \(\sigma\) the variance. The learned orientation field helps the Gabor filter-based enhancement algorithm to enhance the noisy fingerprint image. 

A simplified Convolutional Neural Network (CNN) for Orientation Prediction is shown in Eq. \ref{eq:cnn}
\begin{equation}
\text{Output}(x,y) = \sigma \left( \sum_{i,j} \text{Kernel}(i,j) \times \text{Input}(x-i, y-j) + b \right)
    \label{eq:cnn}
\end{equation}
where \(\sigma\) is an activation function (e.g., ReLU). 

The well-established superiority of the \textit{Convolutional Neural Networks} (CNN) for pattern classification and the recognition problem inspired Cao et al. \cite{cao2015latent} to use it for a challenging problem of estimation of the orientation field of latent fingerprint. 

Schuch et al. \cite{schuch2017convnet} applied CNN trained for regression and achieved an RMSE (root mean square error):

\[
\text{RMSE} = \sqrt{\frac{1}{N} \sum_{i=1}^{N} (\hat{\theta}_i - \theta_i)^2} = 8.53^{\circ}
\]

on a noisy data set. 

Tang et al. \cite{tang2017fingernet} presented a \textit{Deep Convolutional Network} (DCN) that combines the domain knowledge of fingerprint and representation ability of the deep learning algorithm. The proposed technique was able to correctly estimate the fingerprint orientation for both latent and slap fingerprint. 

Qu et al. \cite{gao2018orienet} designed a \textit{Deep Regression Neural Network} (DRNN) that solves the problem of discontinuity at $0^{\circ}$ and the prediction accuracy was high. 

Wu et al. \cite{wu2013svm} treated the fingerprint orientation as a multi-classification problem and applied a \textit{Support Vector Machine} (SVM) to solve it. The SVM decision function is given by:

\[
f(x) = \sum_{i=1}^N \alpha_i K(x_i, x) + b,
\]

where \(K\) is the kernel function. Their experiment shown that the proposed technique is robust and solves the problem of local noise.

\subsubsection{Ridge orientation model learning}
Ram et al. \cite{ram2009active} develop a statistical model called \textit{Active Fingerprint Ridge Orientation Model} (AFROM), which can deform according to input. They use \textit{Legendre Polynomials} to represent the orientation field of fingerprint in the model:

\[
\theta(x, y) = \sum_{n=0}^{N} \sum_{m=0}^{M} a_{nm} P_n(x) P_m(y),
\]

where \(P_n(x)\) and \(P_m(y)\) are Legendre polynomials of degree \(n\) and \(m\), and \(a_{nm}\) are the model coefficients. The AFROM model can estimate the orientation field of a noisy fingerprint image and also can be used for interpolation and extrapolation. The deformation of the model to fit observed data can be formulated as:

\[
\min_{a_{nm}} \sum_{(x,y) \in \Omega} \left( \theta_{\text{obs}}(x,y) - \theta(x,y) \right)^2,
\]

where \(\theta_{\text{obs}}(x,y)\) is the observed noisy orientation and \(\Omega\) is the fingerprint region.

Zhang et al. \cite{zhang2014overlapped} proposed an \textit{Adaptive Orientation Model Fitting} algorithm for the estimation of orientation field for overlapped latent fingerprints. The fitting is formulated by minimizing an energy function:

\[
E = \sum_{i} \| \theta_i^{\text{obs}} - \theta_i^{\text{model}} \|^2 + \lambda R(\theta^{\text{model}}),
\]

where \(\theta_i^{\text{obs}}\) is the observed orientation, \(\theta_i^{\text{model}}\) is the fitted model orientation, and \(R(\cdot)\) is a regularization term promoting smoothness. The algorithm was able to separate the orientation of component fingerprints.

This work was further extended in \cite{zhang2014adaptive_1} to use global orientation field models to predict and correct the orientations in overlapping areas. The global orientation prior can be expressed as:

\[
\theta^{\text{global}}(x,y) = f_{\text{class}}(x,y;\, \mathbf{p}),
\]

where \(f_{\text{class}}\) is a class-specific orientation model parameterized by \(\mathbf{p}\). The correction combines local and global cues:

\[
\theta^{\text{corrected}}(x,y) = \alpha \, \theta^{\text{local}}(x,y) + (1 - \alpha) \, \theta^{\text{global}}(x,y),
\]

where \(\alpha\) balances local estimates and global prediction.

\subsubsection{Dictionary learning}
Inspired by the spelling correction technique used in natural language processing, Feng et al. \cite{feng2012orientation} proposed a fingerprint orientation technique for enhancement of latent fingerprint. They calculate the reference orientation patches from the group of true orientation field and stored in a dictionary. For the estimation of orientation of a fingerprint first orientation of fingerprint estimated using a local orientation technique. Next the estimated orientation field is divided into overlapping patches. For each patch, it's all six neighbors from the patch dictionary taken as candidate patch. From the candidate patch a final patch is selected based on the compatibility constrain. This technique was further improved in \cite{yang2014localized} to incorporate pose estimation of latent fingerprint. For pose estimation a Hough transformation based algorithm was used. 

\subsection*{3.2.3 Dictionary Learning for Fingerprint Orientation Estimation}

Dictionary learning approaches for fingerprint orientation field estimation draw inspiration from analogous techniques in natural language processing, such as spelling correction. In these methods, a \emph{dictionary} of known orientation patches serves as a reference for correcting or refining an estimated orientation field.

\subsubsection*{1. Reference Dictionary Construction}

Following Feng et al.~\cite{feng2012orientation}, the method begins by collecting a set of \emph{true} orientation fields from high-quality fingerprint images. These orientation fields are divided into overlapping patches:

\[
\mathcal{P} = \{ P_1, P_2, \ldots, P_M \},
\]

where \(P_i \in \mathbb{R}^{w \times h}\) represents the orientation angles in a patch of width \(w\) and height \(h\).

A dictionary \(\mathcal{D}\) is constructed:

\[
\mathcal{D} = \{ D_1, D_2, \ldots, D_K \},
\]

where each \(D_k\) is a reference orientation patch sampled from \(\mathcal{P}\). This dictionary encodes typical orientation patterns seen in clean fingerprint images.

\subsubsection*{2. Local Orientation Estimation}

Given a latent fingerprint image \(I\), an initial coarse orientation field \(\hat{O}\) is estimated using local-gradient-based or filter-based techniques:

\[
\hat{O}(x,y) = \arctan\left(\frac{G_y}{G_x}\right),
\]

where \(G_x, G_y\) are local gradients.

\subsubsection*{3. Patch-Based Matching}

The estimated orientation field \(\hat{O}\) is divided into overlapping patches:

\[
\hat{P}_j \in \mathbb{R}^{w \times h}, \quad j = 1, 2, \ldots, N.
\]

For each query patch \(\hat{P}_j\), candidate patches are retrieved from the dictionary \(\mathcal{D}\) based on their spatial neighborhood:

\[
\mathcal{C}_j = \{ D_{k_1}, D_{k_2}, \ldots, D_{k_6} \},
\]

representing the six nearest neighbors in the dictionary space.

\subsubsection*{4. Compatibility-Based Selection}

A final corrected patch \(D^*_j\) is selected from the candidate set \(\mathcal{C}_j\) by minimizing a compatibility cost function:

\[
D^*_j = \arg\min_{D \in \mathcal{C}_j} \; \mathcal{E}(\hat{P}_j, D),
\]

where \(\mathcal{E}\) is a compatibility metric (e.g., sum of squared differences or angular deviation):

\[
\mathcal{E}(\hat{P}_j, D) = \sum_{(x,y) \in \text{patch}} \left| \hat{O}_j(x,y) - D(x,y) \right|^2.
\]

The corrected orientation field is reconstructed by replacing \(\hat{P}_j\) with \(D^*_j\).

\subsubsection*{5. Incorporating Pose Estimation}

Feng et al.~\cite{Feng2013} improved the original method by adding a \emph{pose estimation} step to align the dictionary patches with the latent fingerprint’s global orientation. This was achieved via a Hough transform-based approach that estimates global rotation \(\theta^*\):

\[
\theta^* = \arg\max_{\theta} \; H(\theta),
\]

where \(H(\theta)\) is the accumulator function in Hough space. All candidate patches are rotated accordingly to match the estimated global pose.

\subsubsection*{6. Multi-Scale Dictionary Learning}

Chen et al.~\cite{chen2015multiscale} extended this concept to a multi-scale framework, observing that:

- Small-scale dictionaries have high accuracy for clean regions.  
- Large-scale dictionaries are more robust to noise.

They defined dictionaries at multiple scales:

\[
\mathcal{D}^{(s)} = \{ D^{(s)}_1, D^{(s)}_2, \ldots \}, \quad s \in \{1, 2, \ldots, S\}.
\]

The orientation field is estimated hierarchically across scales, integrating estimates:

\[
\hat{O}(x,y) = \sum_{s=1}^{S} w_s \hat{O}^{(s)}(x,y),
\]

where \(w_s\) are scale weights.

They further formulated the problem using a multi-layer Markov Random Field (MRF) model:

\[
P(\mathbf{O}) \propto \exp\left( - \sum_{c \in \mathcal{C}} \psi_c(\mathbf{O}_c) \right),
\]

where \(\mathcal{C}\) denotes cliques in the MRF and \(\psi_c\) are potential functions enforcing smoothness and compatibility across scales.

\subsubsection*{7. Ridge Structure Dictionary for Segmentation and Enhancement}

Cao et al.~\cite{cao2015latent} extended dictionary-learning ideas to latent fingerprint \emph{segmentation} and \emph{enhancement}. They proposed learning dictionaries of \emph{ridge structures} rather than just orientation fields. Given an input patch \(I_p\), they searched a ridge-structure dictionary:

\[
\mathcal{R} = \{ R_1, R_2, \ldots, R_L \},
\]

to find the best match:

\[
R^* = \arg\min_{R \in \mathcal{R}} \; \| I_p - R \|^2,
\]

which enables improved segmentation masks and frequency field estimates alongside enhanced orientation estimation.

The accuracy of the small-scale dictionary is high, but a large-scale dictionary is more robust against noise in fingerprint. Based on this knowledge Chen et al. \cite{chen2016multi} extended the original work \cite{feng2012orientation} to a multi-scale version. They integrated the estimated orientation at multi-scale to get a better estimated orientation field. They applied multi-layer \textit{Markov Random Field} (MRF) model for the formulation and solving the problem. The method proposed in \cite{feng2012orientation} was not good enough to work for fingerprint segmentation and frequency field estimation. To solve this problem Cao et al. \cite{cao2014segmentation} proposed a ridge structure dictionary-learning based technique for segmentation and enhancement of latent fingerprint.

Dictionary learning methods iteratively refine latent fingerprint orientation fields by leveraging libraries of known orientation or ridge patterns. These techniques incorporate patch-based matching, pose estimation, multi-scale modeling, and probabilistic frameworks (e.g., MRFs) to improve robustness against noise and image degradation.

\subsubsection{Markov Model} 
A Fingerprint orientation estimation technique based on smoothing of the local ridge structure of fingerprint performance is satisfactory for good quality image, bur very poor in low quality region of the image. To solve this problem, Kuang-chih Lee and Salil Prabhakar proposed an orientation estimation technique based on probabilistic technique \cite{lee2008probabilistic}. They inferred the orientation field from the constructed Markov Random Field (MRF). The proposed MRF composed of two parts. The first part model the global orientation of the fingerprint and the second part enforce the smoothing in local regions. 

\subsection*{Probabilistic Orientation Estimation using Markov Random Fields}

Traditional fingerprint orientation estimation methods often rely on smoothing local ridge structures. While these techniques perform well on high-quality images, their performance degrades severely in noisy or low-contrast regions. To address this limitation, Lee and Prabhakar 
proposed a probabilistic approach that formulates orientation field estimation as an inference problem on a Markov Random Field (MRF).

\subsubsection*{1. Problem Definition}

Let \( I \) denote the fingerprint image, and let the continuous orientation field be represented as:

\[
\mathbf{O} = \{ O_s \}_{s \in \mathcal{S}},
\]

where \( O_s \) is the orientation angle at pixel or block location \( s \), and \(\mathcal{S}\) is the set of all sites (pixels or blocks) in the image.

The goal is to estimate \(\mathbf{O}\) given noisy local measurements while enforcing both global and local consistency.

\subsubsection*{2. MRF Model Formulation}

The fingerprint orientation field is modeled as a Markov Random Field with the following joint probability distribution:

\[
P(\mathbf{O} \mid I) \propto \exp\left( - E(\mathbf{O}, I) \right),
\]

where \( E(\mathbf{O}, I) \) is an energy function capturing both data fidelity and smoothness priors.

\subsubsection*{3. Energy Function Components}

The total energy \(E\) comprises two key terms:

\[
E(\mathbf{O}, I) = E_{\text{global}}(\mathbf{O}) + E_{\text{local}}(\mathbf{O}, I).
\]

\paragraph{(a) Global Orientation Model}

The global term enforces overall fingerprint ridge flow consistency, modeling the large-scale, smooth trend of the orientation field:

\[
E_{\text{global}}(\mathbf{O}) = \sum_{s \in \mathcal{S}} \lambda_g \left| O_s - \mu_s \right|^2,
\]

where:
- \(\mu_s\) is the global orientation trend (e.g., derived from fingerprint class models like loops or whorls).
- \(\lambda_g\) controls the weight of global consistency.

\paragraph{(b) Local Smoothing Model}

The local term enforces smoothness between neighboring orientations, reducing noise:

\[
E_{\text{local}}(\mathbf{O}, I) = \sum_{(s, t) \in \mathcal{N}} \lambda_l \, \phi(O_s, O_t),
\]

where:
- \(\mathcal{N}\) is the set of neighboring site pairs.
- \(\phi(O_s, O_t)\) is a smoothness potential, often defined as:

\[
\phi(O_s, O_t) = \left| O_s - O_t \right|^2.
\]

- \(\lambda_l\) controls the strength of local smoothing.

\subsubsection*{4. MAP Estimation}

The optimal orientation field \(\hat{\mathbf{O}}\) is obtained by maximizing the posterior probability, equivalently minimizing the total energy:

\[
\hat{\mathbf{O}} = \arg\min_{\mathbf{O}} \; E_{\text{global}}(\mathbf{O}) + E_{\text{local}}(\mathbf{O}, I).
\]

\subsubsection*{5. Inference Method}

Lee and Prabhakar used iterative optimization techniques (e.g., Iterated Conditional Modes or other local search methods) to solve:

\[
O_s^{(k+1)} = \arg\min_{O_s} \left[ \lambda_g (O_s - \mu_s)^2 + \lambda_l \sum_{t \in \mathcal{N}(s)} (O_s - O_t^{(k)})^2 \right].
\]

At each step, the orientation at site \(s\) is updated based on the global model and neighboring orientations.

\subsubsection*{6. Summary of the Model}

- **Global model** encodes the coarse, ridge-flow pattern typical of the entire fingerprint.  
- **Local smoothing** enforces spatial consistency, reducing noise in the orientation field.  
- **MRF formulation** provides a principled probabilistic framework, balancing prior knowledge with observed local estimates.  

This probabilistic approach improves robustness in noisy or low-quality regions, outperforming naive local smoothing methods.

\subsubsection{Hybrid}
Mardia et al. \cite{mardia1997statistical} first calculate the orientation filed from semi-variogram. This orientation field is quite noisy, local smoothing is required. For smoothing the orientation image, it is first converted to directed vector, then it is processed by a $ 3\times 3$ averaging filter. The orientation field so obtained still need some enhancement. They proposed a technique based on Bayesian framework for the enhancement of the estimated orientation filed.
Turroni et al. \cite{turroni2011improving} combine the best part of local analysis of orientation and machine learning based global technique to propose an improved fingerprint orientation extraction technique that outperforms the most of the technique on challenging dataset.

\subsection{Model Fitting based}
Sherlock and Monro presented a mathematical model for computation of the \textit{Local Ridge Orientation} (LRO) from core and delta positions \cite{sherlock1993model}. Their model computes the orientation field at location \((x,y)\) as:

\[
\theta(x,y) = \arctan\left( \frac{y - y_{\text{core}}}{x - x_{\text{core}}} \right) - \arctan\left( \frac{y - y_{\delta}}{x - x_{\delta}} \right),
\]

where \((x_{\text{core}}, y_{\text{core}})\) and \((x_{\delta}, y_{\delta})\) are the core and delta positions, respectively. Their model has an intelligent tool to resolve the ambiguities in orientation. However, the estimated orientation field differs from the actual orientation field and has limited practical use. 

To solve this problem, a nonlinear model was proposed by Vizcaya and Gerhardt \cite{vizcaya1996nonlinear}. They introduced a correction term to better fit observed orientation:

\[
\theta(x,y) = \theta_{\text{linear}}(x,y) + \sum_{k} \lambda_k \phi_k(x,y),
\]

where \(\theta_{\text{linear}}(x,y)\) is the initial linear field, \(\phi_k\) are basis functions, and \(\lambda_k\) are learned weights. They compared the two models to prove the advantages of their model. 

A fingerprint orientation model suitable for all types of fingerprints was proposed by Zhao and Gu \cite{zhou2004modeling} using rational complex functions:

\[
z = x + i y, \quad f(z) = \frac{P(z)}{Q(z)},
\]

where \(f(z)\) encodes the complex orientation field, and \(P(z), Q(z)\) are polynomials in \(z\). 

Gu et al. \cite{gu2004combination} proposed the polynomial model for the global orientation field:

\[
\theta_{\text{global}}(x,y) = \sum_{n,m} a_{nm} x^n y^m,
\]

and a point-charge model for estimation of the orientation field at singularity points:

\[
\theta_{\text{singularity}}(x,y) = \sum_{k} q_k \arctan\left( \frac{y - y_k}{x - x_k} \right),
\]

where \(q_k\) is the charge strength at singularity \((x_k, y_k)\). They combined these two models using a weighted function to get more robust and accurate fingerprint orientation estimation:

\[
\theta_{\text{combined}}(x,y) = w(x,y) \theta_{\text{global}}(x,y) + (1 - w(x,y)) \theta_{\text{singularity}}(x,y),
\]

where \(w(x,y)\) is a smooth weighting function. Zhou and Gu further improved their work by enforcing smoothness except at several singular points \cite{zhou2004model}.

An algorithm was proposed by Li et al. for modeling the fingerprint orientation field \cite{li2006constrained}. The proposed algorithm comprises two steps. In the first step, orientation was predicted for those areas of the fingerprint image where ridge information was not available or the coherence of the orientation field was low using a piece-wise phase partial model:

\[
\theta(x,y) = \sum_{k} \theta_k \chi_k(x,y),
\]

where \(\chi_k(x,y)\) are indicator functions for regions. For the second step, a constrained nonlinear phase partial algorithm was proposed:

\[
\min_{\theta} \int_{\Omega} \left| \nabla \theta - \mathbf{v} \right|^2 \, dx \, dy + \lambda \, C(\theta),
\]

where \(\mathbf{v}\) is the observed phase gradient and \(C(\theta)\) enforces the singularity structure. The performance of the proposed technique was tested on the NIST-4 database. It fails only for those fingerprint images where the orientation information cannot be extracted reasonably clearly. This technique was analyzed in \cite{li2007stability} for stability. The constrained nonlinear phase partial model \cite{li2006constrained} fails if two singular points are close. A new model was proposed to deal with this problem \cite{li2007stability}. The new model has only one constraint which keeps the singularity at the singular points and abandons the $1^{\text{st}}$ order phase partial constraints:

\[
C(\theta) = \sum_{k} \left| \theta(s_k) - \theta_k^* \right|^2,
\]

where \(s_k\) is the singular point location and \(\theta_k^*\) is its prescribed orientation.

Based on quadratic differentials, a global model for the fingerprint orientation field was proposed in \cite{huckemann2008global}. For three classes (arches, loops and whorls) different models were designed using quadratic differentials:

\[
\omega(z) = \phi(z) \, dz^2,
\]

where \(\phi(z)\) is a meromorphic function describing the fingerprint class. Parameters used in the model are invariant under Euclidean motions. The model allows extrapolation into unobserved regions of fingerprint images. This work was further extended in \cite{gottschlich2016perfect}. The locally adaptive methods are combined to get more robust global models. 

Wang et al. proposed a \textit{Fingerprint Orientation Model based on 2D Fourier Expansions} (FOMFE) that does not require prior knowledge of singular points \cite{wang2007fingerprint}. The \textbf{\textit{Fingerprint Orientation Model based on 2D Fourier Expansion (FOMFE)}}, introduced by Wang et al. in 2007 \cite{wang2007fingerprint}, offers a robust method for modeling the local ridge orientation in fingerprint images using a \textit{2D Fourier series expansion}. This approach is particularly advantageous as it \textit{does not require prior knowledge of singular points (SPs)} and can effectively describe the overall ridge topology, including SP regions, even in the presence of noise. 

For a bivariate function \( f(x, y) \) defined in a rectangular region \( \mathcal{R} = [ -l, l ] \times [ -h, h ] \), the 2D Fourier expansion is expressed as (Eq. \ref{eq:2dfe}):

\begin{align}
f(x, y) = \sum_{m=0}^{k} \sum_{n=0}^{k} \varphi_{mn} \Big[ 
    &\, a_{mn} \cos(m \omega_x x) \cos(n \omega_y y) 
     + b_{mn} \sin(m \omega_x x) \cos(n \omega_y y) \notag \\
    &+ c_{mn} \cos(m \omega_x x) \sin(n \omega_y y) 
     + d_{mn} \sin(m \omega_x x) \sin(n \omega_y y) 
\Big]
\label{eq:2dfe}
\end{align}
where:
\( \omega_x = \frac{\pi}{l} \) and \( \omega_y = \frac{\pi}{h} \) are the fundamental frequencies along the \( x \)- and \( y \)-axes, respectively.
\( a_{mn}, b_{mn}, c_{mn}, d_{mn} \) are the Fourier coefficients to be estimated.
\( \varphi_{mn} \) is a scaling factor defined as (Eq. \ref{eq:sf}):
\begin{equation}
  \varphi_{mn} =
  \begin{cases}
    \frac{1}{4} & \text{if } m = n = 0 \\
    \frac{1}{2} & \text{if } m > 0, n = 0 \text{ or } m = 0, n > 0 \\
    1 & \text{if } m > 0, n > 0
  \end{cases}
\label{eq:sf}
\end{equation}
In the context of fingerprint orientation modeling, the FOMFE decomposes the local ridge orientation into a series of sinusoidal basis functions. The coefficients \( a_{mn}, b_{mn}, c_{mn}, d_{mn} \) capture the periodic variations in ridge orientation across the fingerprint image. By estimating these coefficients, the model can reconstruct the local orientation field, which is essential for tasks such as fingerprint enhancement, singular point detection, and indexing.


    

For partial fingerprint identification, an analytical approach for global ridge orientation modeling based on the inverse orientation model was presented that improves the retrieval rate significantly \cite{wang2010global}. The inverse orientation model can be expressed as:

\[
\theta(x,y) = \arctan\left( \frac{y - y_s}{x - x_s} \right)^{-1},
\]

where \((x_s, y_s)\) is a singularity location, allowing estimation of orientation by inverting the field model. 

Tashk et al. modified the FOMFE model using the Coherence Matrix for fingerprint ridge orientation estimation \cite{tashk2009improvement}. The Coherence Matrix \( \mathbf{C} \) in a local block is defined as:

\[
\mathbf{C} = 
\begin{bmatrix}
\langle G_x^2 \rangle & \langle G_x G_y \rangle \\
\langle G_x G_y \rangle & \langle G_y^2 \rangle
\end{bmatrix},
\]

where \(G_x, G_y\) are gradient components, and \(\langle \cdot \rangle\) denotes local averaging. The coherence value is:

\[
\mu = \frac{\lambda_1 - \lambda_2}{\lambda_1 + \lambda_2},
\]

where \(\lambda_1, \lambda_2\) are eigenvalues of \(\mathbf{C}\). This coherence guides weighting in FOMFE to improve accuracy.

Tao et al. \cite{tao2010estimation} applied \textit{Harris Corner Strength} (HCS) for orientation field estimation to remove abrupt changes in the orientation field. The Harris Corner Strength is calculated as:

\[
\text{HCS} = \det(\mathbf{C}) - k \, \text{trace}^2(\mathbf{C}),
\]

where \(k\) is a sensitivity parameter. The normalized HCS was used as a weight in the FOMFE method to propose weighted 2D Fourier expansion (W-FOMFE):

\[
f(x,y) = \sum_{m,n} w_{mn} \, \Phi_{mn}(x,y),
\]

where \(w_{mn}\) incorporates HCS-derived weights.

Using Legendre polynomials, a curvature-preserving fingerprint ridge orientation smoothing technique was proposed in \cite{ram2008curvature, ram2010modelling}. Their smoothing technique approximates the orientation field as:

\[
\theta(x,y) = \sum_{n=0}^{N} \sum_{m=0}^{M} a_{nm} P_n(x) P_m(y),
\]

where \(P_n\) are Legendre polynomials and \(a_{nm}\) are coefficients optimized for curvature preservation.

Their smoothing technique was based upon the orientation smoothing method proposed by Witkin and Kass \cite{kass1987analyzing}, which penalizes gradient magnitude variations:

\[
E = \int_{\Omega} \left| \nabla \theta \right|^2 \, dx \, dy.
\]

Tashk et al. used filtering and model-based orientation smoothing \cite{tashk2010chebyshev_1}. For filtering, they applied a Gaussian filter:

\[
\theta_{\text{filtered}}(x,y) = G_{\sigma} * \theta(x,y),
\]

where \(G_{\sigma}\) is a Gaussian kernel. For smoothing, they approximated with Legendre or Chebyshev Type I or II polynomials:

\[
\theta(x,y) = \sum_{n,m} a_{nm} T_n(x) T_m(y),
\]

where \(T_n\) are Chebyshev polynomials.

Their method does not require information of singularity points. Jirachawenget et al. \cite{jirachaweng2011residual} first reconstructed the orientation field using a lower-order Legendre polynomial to get the global orientation:

\[
\theta_{\text{global}}(x,y) = \sum_{n,m}^{\text{low}} a_{nm} P_n(x) P_m(y),
\]

then dynamically refined it in singularity regions using higher-order Legendre polynomials:

\[
\theta_{\text{singularity}}(x,y) = \sum_{n,m}^{\text{high}} b_{nm} P_n(x) P_m(y).
\]

The fingerprint orientation estimation algorithm needs to perform well in the noisy region in smoothing the orientation field and also preserve the orientation in the neighborhood of the singularity. Based on weighted \textit{Discrete Cosine Transform} (DCT), Liu et al. \cite{liu2014fingerprint} proposed a fingerprint orientation field reconstruction algorithm:

\[
\theta(x,y) = \sum_{u,v} w_{uv} \cos\left(\frac{u \pi x}{M}\right) \cos\left(\frac{v \pi y}{N}\right),
\]

where \(w_{uv}\) are weights adapted to image quality, achieving good performance in both smooth and singularity regions.

Bian et al. \cite{bian2014fingerprint} used a technique based on quadratic approximation by orthogonal polynomials in two discrete variables:

\[
\theta(x,y) \approx \sum_{i,j} c_{ij} \phi_i(x) \psi_j(y),
\]

where \(\phi_i, \psi_j\) are orthogonal basis functions. The \textit{Linear Projection Analysis} (LPA) is used for estimation of the orientation of local regions:

\[
\hat{\theta}_{\text{LPA}} = \mathbf{W}^T \mathbf{x},
\]

where \(\mathbf{W}\) is the learned projection matrix and \(\mathbf{x}\) is the local feature vector.

Based on sparse coding and discrete cosine transform, Lui and Lui \cite{liu2012fingerprint} proposed a fingerprint orientation field model:

\[
\theta(x,y) \approx \sum_{k} \alpha_k \, \text{DCT}_k(x,y),
\]

where \(\alpha_k\) are sparse coefficients and \(\text{DCT}_k\) are DCT basis atoms.

Gupta and Gupta \cite{gupta2015fingerprint, gupta2016accurate} proposed a fingerprint orientation field algorithm using a model-based technique. The model used in their algorithm is based on the weighted Legendre basis:

\[
\theta(x,y) = \sum_{n,m} w_{nm}(x,y) \, a_{nm} P_n(x) P_m(y),
\]

where weights \(w_{nm}(x,y)\) are computed to ensure good performance in both noisy regions and in the neighborhood of singularity points. They applied three conditions on calculation of weight using symmetric filters required for modeling: (i) for preserving the true orientation in the neighborhood of singularity points, high weight should be assigned in these regions; (ii) in regions of bad quality image (due to scar marks, bruises, sensor noise or finger condition (dry/wet)) should be assigned low weight; and (iii) in regions of good quality image (uniform ridge-valley flow) should be assigned high weight.

\begin{landscape}
\begin{longtable}{
|>{\raggedright\arraybackslash}p{3cm}|
>{\raggedright\arraybackslash}p{2.5cm}|
>{\raggedright\arraybackslash}p{2.5cm}|
>{\raggedright\arraybackslash}p{2.5cm}|
>{\raggedright\arraybackslash}p{2cm}|
>{\raggedright\arraybackslash}p{3cm}|
>{\raggedright\arraybackslash}p{3cm}|
}
\caption{Performance Comparison of Fingerprint Orientation and Enhancement Methods}%
\label{tab:Performance_Comparison_Table} \\

\hline
\textbf{Author} & \textbf{Method Type} & \textbf{Technique} & \textbf{Application} & \textbf{Performance} & \textbf{Advantages} & \textbf{Limitations} \\
\hline
\endfirsthead

\multicolumn{7}{c}{{\tablename\ \thetable{} -- continued from previous page}} \\
\hline
\textbf{Author} & \textbf{Method Type} & \textbf{Technique} & \textbf{Application} & \textbf{Performance} & \textbf{Advantages} & \textbf{Limitations} \\
\hline
\endhead

\hline \multicolumn{7}{r}{{Continued on next page}} \\
\endfoot

\hline
\endlastfoot

Rhodes (1956)\cite{rhodes1956} & Classical & Manual study & Classification & Very low & First foundation & Outdated \\
Scott (1951)\cite{scott1951} & Classical & Manual feature study & Mechanics & Very low & Early insight & Not automated \\
Maltoni et al. (2009)\cite{maltoni2009handbook} & Classical & Review & Recognition & High & Wide coverage & No deep learning \\
Feng \& Jain (2010)\cite{feng2010fingerprint} & Model-based & Phase Reconstruction & Reconstruction & Very high & Sparse minutiae input & Noise sensitive \\ 
Trivedi (2018)\cite{trivedi2018secure} & Secure template & Non-invertible & Matching & High & Privacy preserving & Complex model \\
Mehtre (1987)\cite{mehtre1987segmentation} & Model-based & Direction segmentation & Enhancement & High & Noisy image support & Assumes local uniformity \\
Ratha (1995)\cite{ratha1995adaptive} & Model-based & Adaptive flow & Feature extraction & High & Distortion handling & Costly \\
Bazen (2002)\cite{bazen2002systematic} & Model-based & Direction field & Orientation & Very high & Systematic design & Sensitive singularities \\
Oliveira (2008)\cite{oliveira2008multiscale} & Model-based & Morphological tools & Ridge Repair & High & Broken ridge fixing & Artifacts possible \\
Wang (2007)\cite{wang2007fingerprint} & Model-based & Gradient field & Orientation & High & Simple, effective & Weak at core points \\
Ji \& Yi (2008)\cite{ji2008novel} & Model-based & Ridge projection & Orientation & High & Noise tolerant & Needs tuning \\
Wu et al. (2013)\cite{wu2013svm} & ML & SVM & Orientation Estimation & High & Trainable & Requires data \\
Sahasrabudhe (2013)\cite{sahasrabudhe2013learning} & ML & Continuous RBM & Orientation & Very High & Learns ridge flow & High training cost \\
Cao \& Jain (2015)\cite{cao2015latent} & Deep Learning & CNN & Orientation & Very High & Strong on latents & Data dependent \\
Schuch et al. (2017)\cite{schuch2017cnn} & Deep Learning & ConvNet Regression & Orientation & Very High & Smooth output & GPU needed \\
Tang (2017)\cite{tang2017fingernet} & Deep Learning & FingerNet & Minutiae Extraction & Very High & Unified model & Complex design \\
Qu (2018)\cite{qu2018orienet} & Deep Learning & OrieNet & Orientation & Very High & Latent suitable & Overfitting risk \\
Feng et al. (2012)\cite{feng2012orientation} & Probabilistic & Local modeling & Latent enhancement & Very High & Poor image support & Slow inference \\
Yang et al. (2014)\cite{yang2014local} & Sparse Coding & Dictionary & Orientation & Very High & Flexible & Training cost \\
Jain \& Cao (2015)\cite{jain2015template} & Sparse Coding & Ridge dictionaries & Feature extraction & High & Fine detail capture & Dictionary large \\
Huckemann et al. (2008)\cite{huckemann2008model} & Math Model & Quadratic Diff. & Orientation modeling & High & Global modeling & Not robust \\
Gottschlich (2016)\cite{gottschlich2016perfect} & Model-based & Local/global model & Orientation & Very High & Balances precision & Intensive computation \\
\end{longtable}
\end{landscape}

\section{Chronological Insights}
\begin{figure}[!ht]
\centering
\includegraphics[scale=.45]{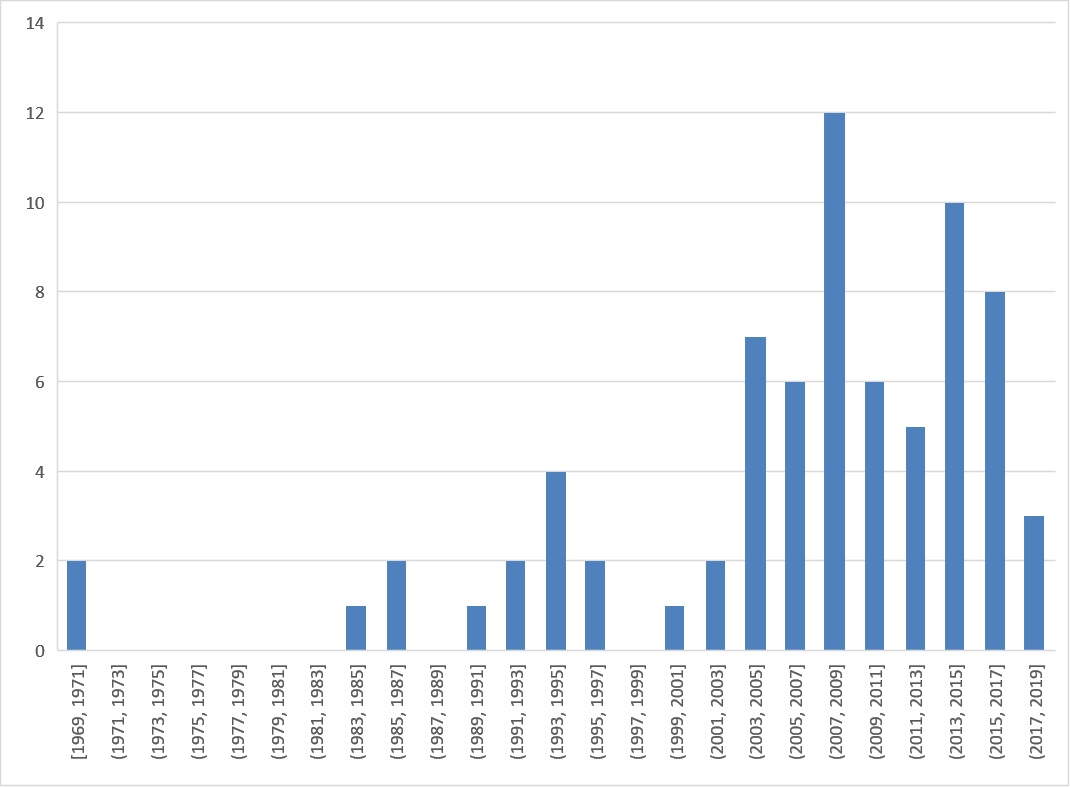}
\caption{Research trend in fingerprint orientation estimation over five Decades: Growth, Peaks, and Stabilization (1969 to 2025)}
\label{fig:hist}
\end{figure}

In this section a timeline-based analysis of the research trends in fingerprint orientation estimation is presented(see Figure \ref{fig:hist}. The histogram in the figure \ref{fig:hist} illustrates the distribution of representative research publications in leading journal and conferences over last five decades. For better visualization, the research trend intervals of two-years is taken. The smaller interval offers better insights into the evaluation and maturation of field over time. The histogram bar in the figure \ref{fig:hist} represents the numbers of publications in the given interval and highlights the phases of activity trend as low activity, growth, peak periods, and eventual stabilization or decline.

\subsection{ Slow and Sporadic Research Activity ($1969 ~to~ 1991$)}

The first two decades, spanning from 1969 to 1991, demonstrate minimal research activity, with only a few intervals showing any significant number of publications. For instance:
\begin{enumerate}

    \item  \textbf{1969–1971:} This interval marks the start of the observed research activity with two publications. Despite this early contribution, subsequent years reflect little to no progress.
    \item  \textbf{1987–1989:} A slight resurgence occurs, with another peak of two publications. 

\end{enumerate}
This period reflects a nascent phase in the research field, likely characterized by limited technological advancements, fewer researchers, and niche interest. Research was sporadic, and substantial attention to the field had not yet materialized.

\subsection{Gradual Growth in Activity ($1993 ~to ~ 2003$)}

The 1990s signal the beginning of a more consistent pattern in research contributions. The intervals 1993–1995 and 1997–1999 show modest increases, with a few more publications recorded in these years compared to earlier decades. 
\begin{enumerate}

\item \textbf{1993–1995:}The slight uptick hints at emerging interest and recognition of the field's potential. 
\item \textbf{2001–2003: }A more noticeable rise begins, suggesting technological advancements and increasing academic attention.
\end{enumerate}
This phase marks the transition from sporadic contributions to a period of steady, though modest, growth.

\subsection{Significant Expansion $(2003 ~to~2009)$}

A turning point is observed starting in the early 2000s. The intervals $2003–2005$ and $2005–2007$ exhibit a marked increase in the number of publications, with the field experiencing exponential growth during this period.
\begin{enumerate}

\item \textbf{2003–2005:} The chart reflects a noticeable surge, indicating that the field has garnered broader interest. Likely, this can be attributed to advancements in computing, algorithms, and data collection methods that made research more feasible and impactful.

\item \textbf{2007–2009:} This interval represents a peak, with nearly 12 publications recorded, a substantial jump compared to prior periods.
\end{enumerate}
The rapid growth during these years could be linked to technological revolutions and increased funding for research. It’s also plausible that interdisciplinary collaboration and applications in real-world problems spurred this surge in contributions.

\subsection{The Golden Era of Research $(2011`to~2017)$}

The most significant phase in the chart is between 2011 and 2017. These years mark the zenith of research activity, with back-to-back high peaks observed across multiple intervals.
\begin{enumerate}

\item \textbf{2013–2015:} Research output reaches its peak, with around 13 publications during this period. This marks the highest level of activity in the dataset, suggesting that the field matured significantly during this time.

\item \textbf{2015–2017:} Another period of high activity, almost comparable to the previous peak, signifies sustained interest and contributions from researchers worldwide.
\end{enumerate}
The robust activity in this period reflects the field's widespread recognition, the emergence of new methodologies, and the availability of advanced tools. Increasing collaboration across academia and industry likely played a role in maintaining the momentum.

\subsection{A Recent Decline$(2017~to~2025)$}

The interval 2017–2025 shows a noticeable dip in the number of publications compared to the previous peaks. Although still active, with a significant number of contributions, the chart reflects a stabilization or slight reduction in research intensity.

Several factors could contribute to this decline:
\begin{enumerate}

  \item  A shift in focus toward new or adjacent research areas.
  \item  A saturation of certain subfields, leading to a slowdown in novel contributions.
  \item Changing funding priorities or emerging technologies competing for academic and industrial attention.
  
\end{enumerate}
While the decline is not drastic, it may indicate that the field is entering a phase of maturity, with growth rates leveling off as it consolidates its achievements.

\subsection{Overall Trends and Implications}

The bar chart encapsulates the trajectory of a research field's lifecycle, from its inception to growth, peak activity, and eventual stabilization. Key insights include:
\begin{enumerate}

    \item Slow Start (1969–1991): The field remained niche and underexplored, with sporadic contributions.
    \item Gradual Growth (1993–2001): The 1990s marked the beginning of consistent research, with technological and academic advances laying the groundwork for future expansion.
    \item Exponential Rise (2003–2009): The early 2000s heralded a significant boom, with research activity accelerating at an unprecedented pace.
    \item Peak Period (2011–2017): The field reached its apex, with substantial contributions marking its "Golden Era."
    \item Stabilization (2017–2025): A slight decline in activity suggests the field is maturing, with research efforts consolidating rather than expanding rapidly.
 
\end{enumerate}

The bar chart is a testament to the dynamic nature of research fields, illustrating how they evolve over time. From its humble beginnings, this field has experienced remarkable growth, driven by technological advancements, interdisciplinary collaboration, and global recognition. While the slight decline in recent years may suggest stabilization, it also highlights opportunities for exploring new directions and applications.

This visualization underscores the importance of studying such trends to understand the lifecycle of research areas, aiding in strategic planning for academics, funding agencies, and policymakers. Whether the field will plateau, experience a resurgence, or transition into new domains remains an open question for the years ahead.

%
%
%
\begin{figure}[!ht]
\centering
\includegraphics[scale=.55]{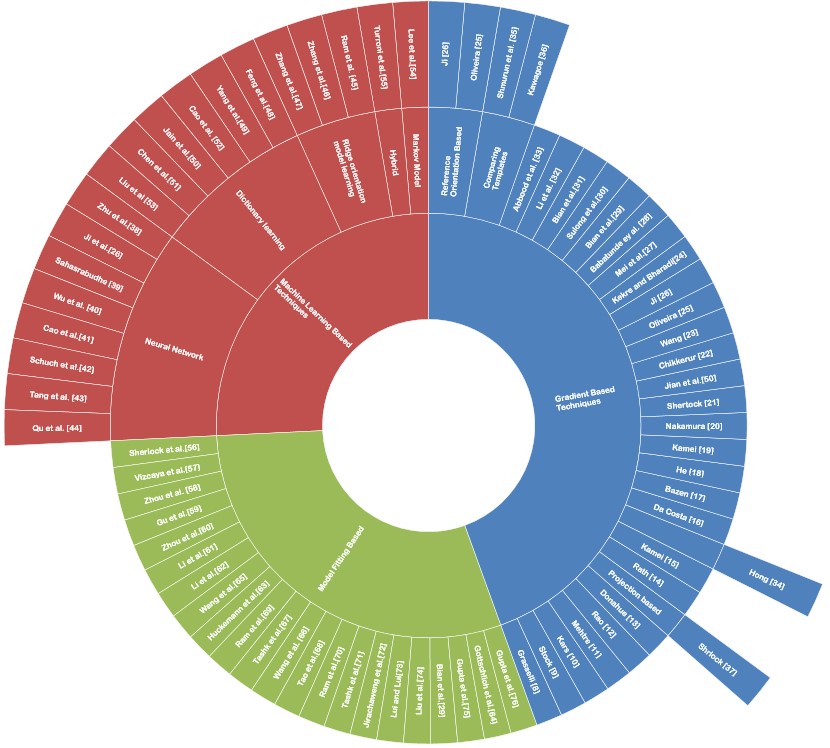}
\caption{Classification of Orientation Estimation Research Approaches in Machine Learning, Gradient-Based Techniques, and Hybrid Methods Across Key Studies and Contributions}
\label{fig:sunbrt}
\end{figure}

The figure \ref{fig:sunbrt}  is a \textit{sunburst chart} that visualizes the relationship between categories, sub-categories, authors, and years of research techniques. The hierarchical structure offers a clear and concise way to represent a dataset categorized under Gradient Based Techniques, Machine Learning Based Techniques, and Model Fitting Based Techniques. The chart has three main categories represented by distinct colors:
\begin{itemize}

 \item Gradient Based Techniques $(Blue)$.
 \item Machine Learning Based Techniques $(Red)$.
 \item Model Fitting Based Techniques $(Green)$.
    
\end{itemize}
Each category is subdivided into sub-categories, which are further divided into the contributions of different authors and their respective years of publication. The hierarchical data flows outward, with the innermost layers being the broad categories and the outermost layers representing individual contributions.

\section{Perform Evaluation of Orientation Estimation}

The performance of fingerprint orientation estimation technique is generally evaluated by measuring the accuracy
of singularity points (Core and delta) detected in an orientation image of the fingerprint. The performance of
some of the techniques has been also reported as accuracy of fingerprint matching and accuracy of orientation
estimation. Some of the very early orientation flow techniques were designed for segmentation of foreground and
background of the fingerprint impression and there was not a quantitative evaluation of the performance.

\subsection*{5.0.1 Fingerprint Matching}

Fingerprint orientation estimation is one of the fundamental steps in feature extraction for most of the fingerprint
biometric systems. So, the accuracy of fingerprint matching is directly affected by the accuracy of orientation
estimation. The False Match Rate (FMR) and False Non-Match Rate (FNMR) are two widely used measurements
to report the performance of fingerprint biometric system.

\paragraph{False Match Rate (FMR):}
\[
\text{FMR}(\tau) = \frac{\text{Number of impostor attempts accepted at threshold } \tau}{\text{Total number of impostor attempts}}.
\]

\paragraph{False Non-Match Rate (FNMR):}
\[
\text{FNMR}(\tau) = \frac{\text{Number of genuine attempts rejected at threshold } \tau}{\text{Total number of genuine attempts}}.
\]

The value where both FMR and FNMR are equal is called the Equal Error Rate (EER), which has been reported in literature as a performance measure for fingerprint orientation estimation algorithms:

\[
\text{EER} = \text{FMR}(\tau^*) = \text{FNMR}(\tau^*),
\]

where \(\tau^*\) is the operating threshold at which the two error rates intersect.

In actual applications, the fingerprint biometric system does not typically operate exactly at the EER. Therefore, other
performance metrics like FMR100 or FMR1000 are generally reported. These are defined as the *lowest* achievable FNMR
values when the system is constrained to very low FMR levels.

\paragraph{FMR100:}
\[
\text{FMR100} = \min_{\tau} \left\{ \text{FNMR}(\tau) \,\big|\, \text{FMR}(\tau) \leq 0.01 \right\}.
\]

\paragraph{FMR1000:}
\[
\text{FMR1000} = \min_{\tau} \left\{ \text{FNMR}(\tau) \,\big|\, \text{FMR}(\tau) \leq 0.001 \right\}.
\]

These metrics indicate the system's robustness when tuned for high-security operating points.

\subsection*{5.0.2 Singularity Point Detection}

The singularity point (SP) is one of the important features of a fingerprint, which is invariant to rotation, translation,
and scaling of the fingerprint. It is generally used for alignment and classification of fingerprints. The Core and
Delta are two very predominant SPs. The Core and Delta singularities lie in regions of abnormality in the orientation
flow. So, the accuracy in detection of singularity points (core or delta) can be used as an indirect measure of the
accuracy of fingerprint orientation flow estimation algorithms.

The performance of orientation flow estimation techniques in terms of Precision, Recall, and F-measure
(Eqs. 10, 11, and 12) are also reported by the respective papers for different standard fingerprint databases:

\paragraph{Precision:}
\[
\text{Precision} = \frac{TP}{TP + FP} \tag{10}
\]
where:
- \( TP \): True Positives (correctly detected SPs)
- \( FP \): False Positives (spuriously detected SPs)

Precision measures the rate of detecting SPs that are the correct ones.

\paragraph{Recall:}
\[
\text{Recall} = \frac{TP}{TP + FN} \tag{11}
\]
where:
- \( FN \): False Negatives (missed SPs)

Recall describes the proportion of correctly detected SPs relative to all true SPs.

\paragraph{F-measure (F1-score):}
\[
\text{F-measure} = \frac{2 \times \text{Recall} \times \text{Precision}}{\text{Recall} + \text{Precision}} \tag{12}
\]

\paragraph{Alternative Form:}
\[
\text{F}_\beta = (1 + \beta^2) \frac{\text{Precision} \times \text{Recall}}{\beta^2 \times \text{Precision} + \text{Recall}}
\]
where \(\beta\) is a parameter controlling the trade-off between Precision and Recall. For \(\beta = 1\), we recover the standard F-measure.

Precision measures the correctness of detected singularities, while Recall measures the completeness. The F-measure balances these two, indicating overall performance. Well-performing orientation flow estimation will be indicated by good SP detection, where higher values of Precision, Recall, and F-measure are achieved.

\section{Challenges and Opportunity}
In Section \ref{or}, we provided a systematic classification of fingerprint orientation flow estimation algorithms, highlighting the salient features of each approach. A clear trend emerges from this review: there has been a pronounced transition from traditional gradient-based and model-fitting techniques toward machine learning–based methods. This shift can be attributed to the improved performance of machine learning approaches, particularly in the context of noisy or partial fingerprint images, where traditional techniques often struggle.

Nevertheless, it is important to recognize that machine learning–based methods typically introduce higher computational complexity. This limitation poses challenges for their deployment in large-scale fingerprint databases and on resource-constrained, portable devices. Furthermore, none of the existing techniques in the literature can be regarded as universally optimal. There remains significant opportunity for the development of fingerprint orientation estimation algorithms that achieve both low computational complexity and high estimation accuracy. Notably, reliable orientation estimation also plays a critical role in applications such as the separation of overlapping latent fingerprints.

Addressing the challenges of fingerprint orientation estimation requires effective strategies to manage poor-quality images, damaged ridge patterns, and pervasive background noise. Contemporary solutions often leverage hybrid methodologies that combine the computational efficiency of gradient-based approaches with the robustness and adaptability of machine learning techniques. In addition, preprocessing procedures—such as contrast enhancement and ridge structure normalization—are widely employed to further enhance the accuracy of orientation estimation.

In conclusion, fingerprint orientation estimation remains a fundamental component in the development of reliable and accurate biometric identification systems. The selection of an appropriate estimation technique must be informed by considerations including input image quality, computational constraints, and specific application requirements. Looking forward, continued advancements in deep learning and hybrid methodologies are anticipated to further improve accuracy and robustness, thereby enabling more reliable performance across a wide range of fingerprint recognition scenarios.

	\section*{Declaration of Generative AI and AI-assisted technologies in the
	 writing process}
	 During the preparation of this work the author(s) used ChatGPT
	 in order to edit the writing of the paper. After using this tool, the
	 author(s) reviewed and edited the content as needed and take(s) full
	 responsibility for the content of the publication.
	
	\bibliography{mybibfile}

\end{document}